\documentclass[a4paper]{article}
\usepackage{todonotes,a4wide}
\usepackage{amsmath,graphicx}
\usepackage{amsfonts,color}
\usepackage{multirow}
\usepackage{algorithm,algcompatible}
\usepackage[footnotesize]{caption} 

\newcommand{\ie}{\emph{i.e.}, }

\newcommand{\eg}{\emph{e.g.}, }

\newcommand{\real}{\mathbb{R}}

\newcommand{\Up}{\bs{\rm Up}}

\DeclareMathOperator*{\argmin}{argmin}

\DeclareMathOperator{\sinc}{\rm sinc}

\DeclareMathOperator{\vecrm}{\rm vec}

\newcommand{\norm}[1]{\left\| #1 \right\|}

\newcommand{\cl}{\mathcal}

\newcommand{\bs}{\boldsymbol}
\newcommand{\bb}{\mathbb}

\def\bx{\bs x}
\def\bal{\bs \alpha}
\def\by{\bs y}
\def\bPhi{\bs \Phi}

\def\Xten{\bs{\mathcal{X}}}
\def\bX{\bs X}
\def\bY{\bs Y}

\def\bu{\bs u}
\def\bxs{\bs x^\star}

\def\bn{\bs n}

\def\bA{\bs A}

\def\st{\ \mathrm{s.t.} \ }
\def\and{\quad \mathrm{and} \quad}

\title{Generalized Inpainting Method\\ for Hyperspectral Image Acquisition}

\author{K. Degraux$^1$, V. Cambareri$^2$, L. Jacques$^1$, B. Geelen$^3$,\\[2mm]
C. Blanch$^3$ and G. Lafruit$^{3,4}$\\[2mm]
\footnotesize
$^1$ISPGroup, ICTEAM, UCLouvain, Belgium.\ $^2$ARCES, University of Bologna, Italy.\\
\footnotesize
$^3$IMEC, Belgium. $^4$ Universit\'e Libre de Bruxelles, Belgium.}

\begin{document}

\maketitle

\begin{abstract}
A recently designed hyperspectral imaging device enables multiplexed acquisition of an entire data volume in a single snapshot thanks to monolithically-integrated spectral filters. Such an agile imaging technique comes at the cost of a reduced spatial resolution and the need for a demosaicing procedure on its interleaved data. In this work, we address both issues and propose an approach inspired by recent developments in compressed sensing and analysis sparse models. We formulate our superresolution and demosaicing task as a 3-D generalized inpainting problem. Interestingly,
the target spatial resolution can be adjusted for mitigating the compression level of our sensing. The reconstruction procedure uses a fast greedy method called Pseudo-inverse IHT.
We also show on simulations that a random arrangement of the spectral filters on the sensor is preferable to regular mosaic layout as it improves the quality of the reconstruction. The efficiency of our technique is demonstrated through numerical experiments on both synthetic and real data as acquired by the snapshot imager.\\[3mm]
\end{abstract}

\section{Introduction}
\label{sec:intro}
Hyperspectral Imaging (HSI) refers to the acquisition and processing of multichannel digital images of the spectrometry of every pixel of a scene, \ie the intensity of light as a function of its wavelength. The knowledge of the spectrum can serve to identify the materials that are present in a particular pixel~\cite{Bioucas12}.
Many scientific applications of this technique exist in, \eg remote sensing~\cite{Shippert2004} and food science~\cite{Gowen2007} with an increasing number of industrial and general purpose applications emerging due to the rapid simplification undergone by recent HSI systems~\cite{Tack2012}.

Hyperspectral images, \ie data volumes (or \emph{cubes}) ordered in two spatial dimensions and one spectral, may contain several millions of voxels, partitioned in anywhere from a dozen up to hundreds of wavelengths.
Traditional HSI systems require either line scanning, \eg with a motorized translation stage, or spectral scanning with tunable filters~\cite{gupta2000progress}. Most often, this means a slow acquisition and a cumbersome device only suitable for imaging in controlled environments. More recently, snapshot HSI was introduced as a trade-off between spatial resolution and acquisition time; this imaging mode is obtained by multiplexing optically the 3-D spatial-spectral information on a 2-D CMOS photosensor array, over which a layer of  Faby-P\'erot (FP) spectral filters is deposited~\cite{Tack2012,geelen2014compact}. The main limitation of this approach is that the spatial resolution of the sensor array actually becomes limited by the number of acquired wavelengths.

Compressed Sensing (CS) counts among the means used to cope with this resolution limit. When the acquisition process is appropriately designed, this mathematical framework leverages the natural structure of a signal (\eg its sparsity) to robustly reconstruct it from only a few non-adaptive measurements~\cite{Candes2006}. Compared to standard 2-D compressive imaging, HSI also offers an extra spectral dimension to exploit. CS is thus clearly an appealing objective for HSI, as noted by some compressive HSI designs~\cite{Gehm2007,Wagadarikar2008a}, some of which tackled explicitly an imaging architecture with a FP-filtered sensor array~\cite{Degraux:iTWIST14}. Nevertheless, these designs often require complex optical elements (\eg spatial light modulation, beam splitter) that increases the global system complexity. 

The contributions of this work are multiple. First, our study is based on the snapshot imager from~\cite{geelen2014compact} and processes the resulting low-resolution data volume with an inpainting method to fill the missing information. The aim is to find a compromise between an instantaneous acquisition without scanning and a reasonably high target resolution with the aid of a recovery algorithm. Second, on the hardware side, we propose to slightly modify a simple snapshot imager design by randomly intermixing the FP filters on the sensor; this avoids aliasing effects due to subsampling on a regular grid. Simulations confirm that a random spatial arrangement of those spectral filters on the sensor is preferable to a regular mosaic layout as it improves HSI quality. Third, on the algorithmic and theoretical side, we generalize the classical inpainting problem by adding a linear interpolation step (\emph{upscaling}) in the forward model enabled by an optical spatial lowpass filtering~\cite{optfilter} common in demosaicing methods. Given a fixed number of observations, this linear operation allows to choose the target resolution and in particular, to reduce the subsampling factor by any desired value, making it easier for the inpainting method to fill the missing information. Our method is akin to CS in the sense that the formulations and goals are identical. The difference resides in the acquisition operation: where CS entails random linear mixtures as compressive measurements, our HSI device simply applies a direct subsampling. This work is also related to super-resolution as it allows to increase the number of pixels from the low multiplexed resolution of the focal plane. The proposed inpainting task relies on a greedy algorithm, the Pseudo-inverse IHT~\cite{Peyre2010}, slightly adapted to our needs for optimizing its convergence. Finally, we demonstrate through numerical simulations the potential of our design concept and reconstruction method for practical applications. 

The paper is structured as follows. The second section presents the acquisition device architecture and formulates the generalized inpainting problem. The third section presents the redundant dictionary used to promote a low-dimensional signal model, followed by a brief analysis of the PIHT algorithm. The fourth section presents the results of numerical simulations on both synthetic and real data.

\section{Hyperspectral Image Inpainting}
\label{sec:HSInpainting}
\begin{figure}[!th]
  \newlength{\lovspl}\setlength{\lovspl}{1mm}
  \newcommand{\lovsp}{\hspace{\lovspl}}
  \centering
  \includegraphics[height=3cm]{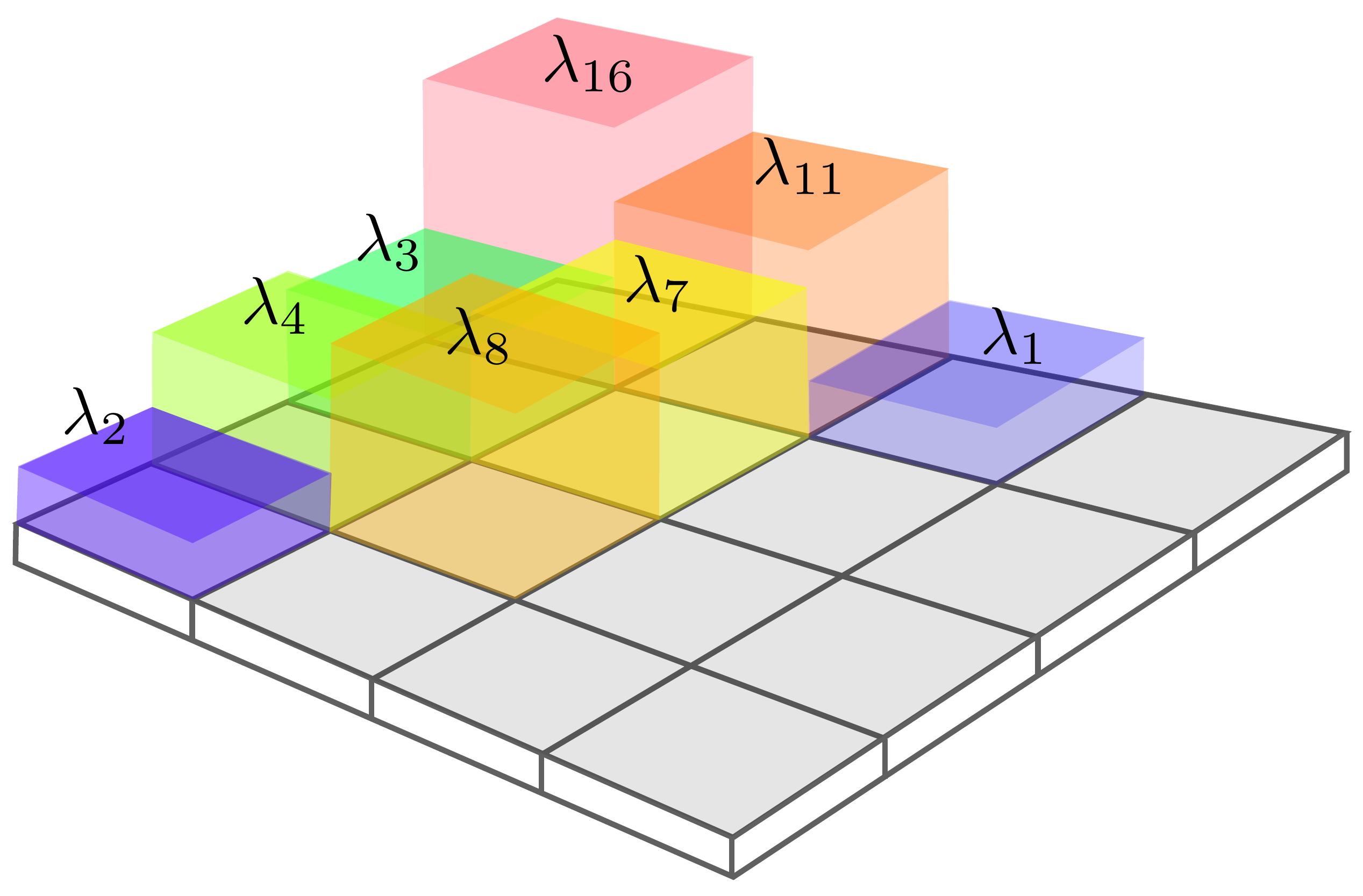}\quad 
  \includegraphics[height=2.5cm]{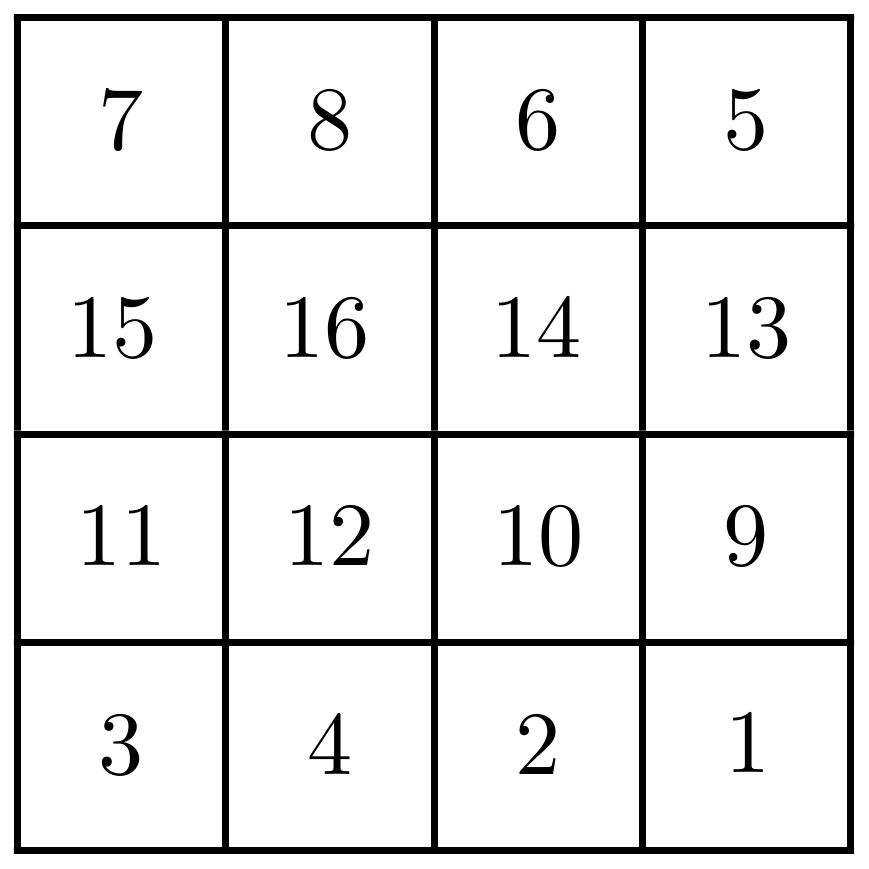}
  \caption{Left: Illustration representing a $4\!\times\!4$ CMOS macropixel integrated with 16 Fabry-P\'erot filters (not all displayed). Right: True wavelength index selection in one such macropixel of the HSI mosaic imager~\cite{geelen2014compact}.}
  \label{fig:FPA-FPI-explan}
 \end{figure}

This work proposes a HSI sensor design on what could be called ``direct domain-compressed sensing", \ie random subsampling. This architecture is associated to a snapshot spectral sensor that monolithically integrates a set of Fabry-P\'erot interferometers (FPI)~\cite{Degraux:iTWIST14} organized as pixel-level filter mosaics on top of a standard, off-the-shelf CMOS sensor, the \emph{focal plane array} (FPA). Each FPI is actually made of a transparent layer sandwiched between two mirrors, the layer width determining the selected wavelength (See Fig.~\ref{fig:FPA-FPI-explan}).  

Let $\hat{\Xten} \in \real^{N_I\times N_J \times L}$ be a hyperspectral (HS) volume of a target scene ($L$ is the number of bands) and $\bs Y \in \mathbb{R}^{m_I \times m_J}$ represent the recording at the \emph{focal plane array} (FPA) with the same spatial resolution as $\hat{\Xten}$ ($N_I = m_I$, $N_J = m_J$). The pixels of  $\bY$ are a \emph{selection} (\eg random) of a group of voxels of the whole volume $\hat{\Xten}$; more precisely, each pixel $y_{i,j}$ of $\bY$ corresponds to a voxel $\hat{x}_{i,j,\lambda}$ of $\hat{\Xten}$ with the same spatial indices $(i,j)$ and a specific wavelength index $\lambda \in [L]:=\{1,\,\cdots,L\}$, \eg selected by a FP filter. 

One emphasis of this paper is to assume the existence of an optical filtering of the spatial frequencies of the volume $\hat{\Xten}$ that ensures that each spectral band is spatially band-limited (\eg with birefringent filtering~\cite{optfilter}). As will become clearer hereafter, this ensures that each spectral band is spatially smooth or band-limited. This practice is common for stabilizing RGB demosaicing~\cite{Keelan2004}. This can be done for instance by slightly defocusing the objective lens of the imager under the assumption that the observed scene is well contained in the optical depth-of-field (DOF). In this context, we can target the reconstruction of a \emph{smaller spatial resolution} and aim to reconstruct an undersampled volume $\Xten \in \real^{n_I\times n_J \times L}$ while imposing $n_I \leq m_I $ and $n_J \leq m_J$. In other  words, the compressive acquisition of the HS volume can be \emph{mitigated} by varying  $N = n_In_J L$. Typically, if $L = 16$, taking $n_I = m_I/2$ and $n_J = m_J/2$ the number of observations reads $M := m_I m_J = 4N/L = N/4$.

Mathematically, this means that there exists an operator $\Up \in \bb R^{n_I n_J \times M}$ so that, in matrix form,
\begin{equation}
	\hat{\bx}_\lambda\ =\ \Up~\bx_\lambda
\end{equation}
where $\bx_\lambda = \vecrm (\bX_\lambda) \in \bb R^{n_I n_J}$ (resp. $\hat{\bx}_\lambda =  \vecrm (\hat{\bX}_\lambda) \in \bb R^M$) is the vectorized form of the spectral band $\bX_\lambda$ (resp. $\hat{\bX}_\lambda$) of $\Xten$ (resp. $\hat{\Xten}$).

To understand the structure of this operator, let us consider it in a simplified situation, \ie in the sampling model of a one dimensional ``monochromatic image'' $f(t) \in L^2(\real)$ on a spatial domain parametrized by $t$. At a low resolution with spatial discretization period $T>0$, the continuous light intensity is acquired as
$$
\textstyle f[n] = \int_{-\infty}^{\infty} f(t) \delta(t-nT) dt = f(nT),
$$
with $n \in \bb Z$ the indices of each pixel. At a finer resolution with sampling period $0<T'<T$ (\eg $T' = T/4$), we would have samples
\begin{equation}
  \label{eq:gk-eq}
  \textstyle g[k]\ =\ \int_{-\infty}^{\infty} f(t) \delta(t-kT') dt\ =\  f(kT'),  
\end{equation}
with $k \in \bb Z$. Let us assume the existence of a generative (interpolative) model of $f$ 
\begin{equation}
\label{eq:generative-model}
\textstyle	f(t) \approx \hat{f}(t) = \sum_{n=-\infty}^{\infty}  f[n] \varphi_n(t).
\end{equation}
To ensure that the model error $f-\hat{f}$ is minimal, the signal $f$ has to be sufficiently smooth: a condition that can be enforced by an optical filtering, as assumed in this paper. In particular, in the case of the Shannon-Whittaker formula with $\varphi_n(t) =  \sinc\left(\frac{t-nT}{T}\right)$, $f = \hat{f}$ if $f$ has no frequencies higher than $1/2T$ (band-limited), otherwise aliasing occurs. Depending on the properties of $f$, other basis functions $\varphi_n$, possibly with finite support, are also possible for minimizing the model error \cite{Mallat1999}. 

Since $\eqref{eq:generative-model}$ allows us to estimate a continuous signal from the samples $f[n]$ at low resolution, assuming a vanishing model error and combining this equation with \eqref{eq:gk-eq} provides
\begin{equation}
\textstyle  g[k]\ =\ \sum_{n=-\infty}^{\infty}  f[n]\,u[k,n],
\end{equation}
with $u[k,n] = \varphi_n(kT')$ defining the entries of our up-sampling operator $\Up$ in this simplified exposition. Rather than the IIR and impractical filters of the Shannon-Whittaker generative model, we use here the 2-D FIR Lanczos filter~\cite{Duchon79}. 

We are now ready to express the general sensing model of our HS sensing. Since there is ideally no spectral dispersion in the device and each FPA cell is sensitive to exactly one wavelength, we can separate the FPA pixels in subsets, according to their wavelength, and consider that each band of the hyperspectral volume $\Xten$ is acquired independently with a different subset of pixels. Let us denote by $\bs{M}_\lambda \in \{0,1\}^{M\times M}$ the diagonal mask operator selecting the FPA pixels of the band $\lambda$. By construction of the FPA, we have $\sum_\lambda \bs M_{\lambda} = \bs I_{M \times M}$ and $\bs M_\lambda \bs M_{\lambda'} = \bs 0$ for $\lambda \neq \lambda'$.

Writing $\bs y = \vecrm(\bs Y)$ and $\bs x = (\bs x^T_{1},\,\cdots,\bs x^T_{L})^T$, and defining the general sensing matrix $\bPhi := \big(\bPhi_{1},\,\cdots,\bPhi_{L}\big)$ with $\bPhi_{\lambda} := \bs M_{\lambda} \Up$, our \textit{generalized} inpainting problem corresponds now to recover the target HS volume $\bs x$ from the possibly noisy observations $\bs y$ in
\begin{equation}
  \label{eq:noisy-forward-model}
  \by = \bPhi \bx + \bn,  
\end{equation}
with $\bn$ an additive noise of bounded energy.

\section{Signal Prior and Reconstruction}
\label{sec:reconstruction}
Two main families of CS reconstruction methods exist for estimating $\bx$ from \eqref{eq:noisy-forward-model}. Convex relaxations of the associated sparsely-regularized (therefore combinatorial) inverse problem, such as the Basis Pursuit DeNoising (BPDN) program, were among the first methods to be used in CS~\cite{Candes2006}. The second family contains greedy algorithms, \eg Iterative Hard Thresholding (IHT)~\cite{blumensath:2009:IHT}, that aim at solving approximately the combinatorial problem. 
Originally, CS theory was only adapted to exploit sparsity of the signal in an orthonormal basis but alternatives have emerged since then.
In compressive HSI,~\cite{Golbabaee2012,Golbabaee2012a,Golbabaee2012n} used structured sparsity, total variation and low rank. In~\cite{Li2012}, the authors proposed to use an adapted dictionary containing the known spectra of each material potentially present in the scene.
Experience from other data restoration problems (\eg denoising, deconvolution) have shown that using \emph{redundant dictionaries} instead of orthonormal bases often lead to dramatic improvement. 
On the website associated with the book~\cite{Peyre2010}, Gabriel Peyr\'e proposed a modified version of IHT using the Pseudo-Inverse of a redundant analysis dictionary after the thresholding step. In this section, we present this algorithm called here PIHT for Pseudo-Inverse IHT. It is related to AIHT~\cite{Giryes2014} and SSIHT~\cite{Giryes2013}, two extensions of IHT for signals respectively in the \emph{cosparse analysis}~\cite{Nam2013}  and in the \emph{sparse synthesis}~\cite{Rauhut2008} models. However, compared to these two algorithms, each iteration of PIHT is faster to compute as efficient algorithms exist for the computing the pseudo-inverse of several redundant transforms. This is explained below after the definition of our signal prior.\\[-3mm]

\noindent{\bf Sparse analysis model:} Since the inverse problem introduced in Sec.~\ref{sec:HSInpainting} is underdetermined ($M\leq N$), prior information about $\Xten$ must be assumed to enforce the solution unicity. In this work, we adopt an \emph{analysis sparse model} for the HS volume. The bands $\bX_\lambda$ of $\Xten$ are first expected to lead to sparse coefficients when analyzed through an Undecimated Discrete Wavelets Transform (UDWT), \ie a shift-invariant, redundant version of the Discrete Wavelet Transform~\cite{Mallat1999,Starck2007a}. Second, the targeted individual pixel spectra are assumed smooth and we select therefore a Discrete Cosine Transform (DCT) basis for ``sparsifying'' these spectra. These two analysis priors are then combined by tensorial product, \ie we form 
$$
	\bA = \bA_{\mathrm{UDWT}} \otimes \bA_{\mathrm{DCT}}
$$
with $\otimes$ the Kronecker product, such that the transformed volume $\bA \bx \in \real^P$  is sparse. The UDWT uses the biorthogonal Daubechies wavelets of length 8 on 3 levels (such that $P = 10N$). Moreover, the UDWT scaling coefficients are further transformed with a 2-D DCT, as those low-frequency UDWT bands should be sparser after such a transform. This solution led to better results than avoiding this additional DCT transform or simply discarding these unsparse bands from the prior. Finally, all reconstruction methods described below are constrained to provide solutions within a realistic intensity range, \ie $\bs x \in \cl R = [0,x_{\max}]^N$ for some known $x_{\max}>0$, or  $\cl P_{\cl R} \bs x = \bs x$ with $\cl P_{\cl R}$ the orthogonal projector on~$\cl R$. 

\medskip
\noindent{\bf Pseudo-Inverse IHT:} We have selected in this work the Pseudo-Inverse IHT (PIHT)~\cite{Peyre2010}, \ie a \emph{greedy} algorithm that aims at approaching the following non-convex problem:
\begin{equation}
\label{eq:IHT_min}
\textstyle	\bxs = \argmin_{\bs u \in \cl R} \norm{\bPhi \bu - \by} \st \cl H_{K}(\bs A \bu)  = \bs A \bu,
\end{equation}
where the \emph{hard thresholding} operator $\cl H_{K}(\bu)$, sets to 0 all but the $K$ largest entries of $\bu$. 

\begin{algorithm}[h!] 
	\caption{Pseudo-Inverse Iterative Hard Thresholding}               
	\begin{algorithmic}[1]
		\STATE{\textbf{input:}\\ 
		$\bx_0$; $n_\mathrm{iter}$; $K^s, \, \forall s\in[S]$;}
		\STATE {\textbf{initialization:}\\ 
		$\bx^1 := \bx_0;$}
		\FOR{$s = 1$ to $S$}
			\STATE{$\tau^s := \norm{\bPhi^*(\by-\bPhi\bx^s)}^2/\norm{\bPhi\bPhi^*(\by-\bPhi\bx^s)}^2$ }
			\STATE{$\bs a^s := \bx^s + \tau^s \bPhi^* \left( \by - \bPhi \bx^s \right)$}
			\STATE{\label{line:pinv} $\bx^{s+1} := \cl P_{\cl R}\, \big[\bA^\dagger \mathcal{H}_{K^s}\left(\bA \bs a^s\right)$\big]}
		\ENDFOR
	\end{algorithmic}
	\label{alg:IHT}
\end{algorithm}

PIHT is described in Alg.~\ref{alg:IHT}. In order to maximize efficiency and speed, PIHT is initialized with a point that is not ``too far" from $\bs x^\star$ in (\ref{eq:IHT_min}), \ie $\bx_0$ is obtained by a standard 3-D linear interpolation of the unknown samples of $\hat{\bs{\cl X}}$ from those sampled by $\bs Y$ and resizing the result to the dimensions of $\bs{\cl X}$. Inspired by~\cite{Bobin2007,Starck2004}, the number of non-zero components $K$ is progressively increased for the hard thresholding during the $S$ iterations. The initial $K^0$ and final $K^{S}$ values are set following a \emph{heuristic}, depending on the statistical distribution of $\bal_0 = \bA \bx_0$ as a proxy for the one of the unknown $\bA \bx$. This distribution being heavy tailed, we took $K^S$ and $K^0$ as the number of coefficients bigger than the ``mean plus $\sigma$'' and the ``mean plus $2.5\sigma$'' of the distribution of $\log |\bal_0|$, respectively.  Notice that the value $\tau^s$ in Alg.~\ref{alg:IHT} is the $\tau$ minimizing $\|\bs y- \bPhi \big(\bx^s + \tau \bPhi^* \left( \by - \bPhi \bx^s \right)\big)\|$. 

Let us conclude this section by mentioning that, compared to AIHT~\cite{Giryes2014} and SSIHT~\cite{Giryes2013}, PIHT is faster to compute at each iteration as fast algorithms exist to estimate the vector-matrix product involving the pseudo-inverse $\bs A^\dagger$ at line \ref{line:pinv} in Alg.~\ref{alg:IHT}. Those methods rely on the knowledge of the dual transform of our UDWT system~\cite{Mallat1999}. However, convergence guarantees for PIHT are still unknown. We can just state that, if $\bx \in \cal R$ is such that $\by = \bPhi \bx$ (noiseless sensing), then, $\bx$ is a fixed point of the PIHT iterations if and only if $\bs A \bs x$ is $K$-sparse. For space reason, we omit here the proof of this fact.

\section{Numerical Results}
\label{sec:experiments}

\begin{figure}[t]
  \centering
  \includegraphics[width=4cm]{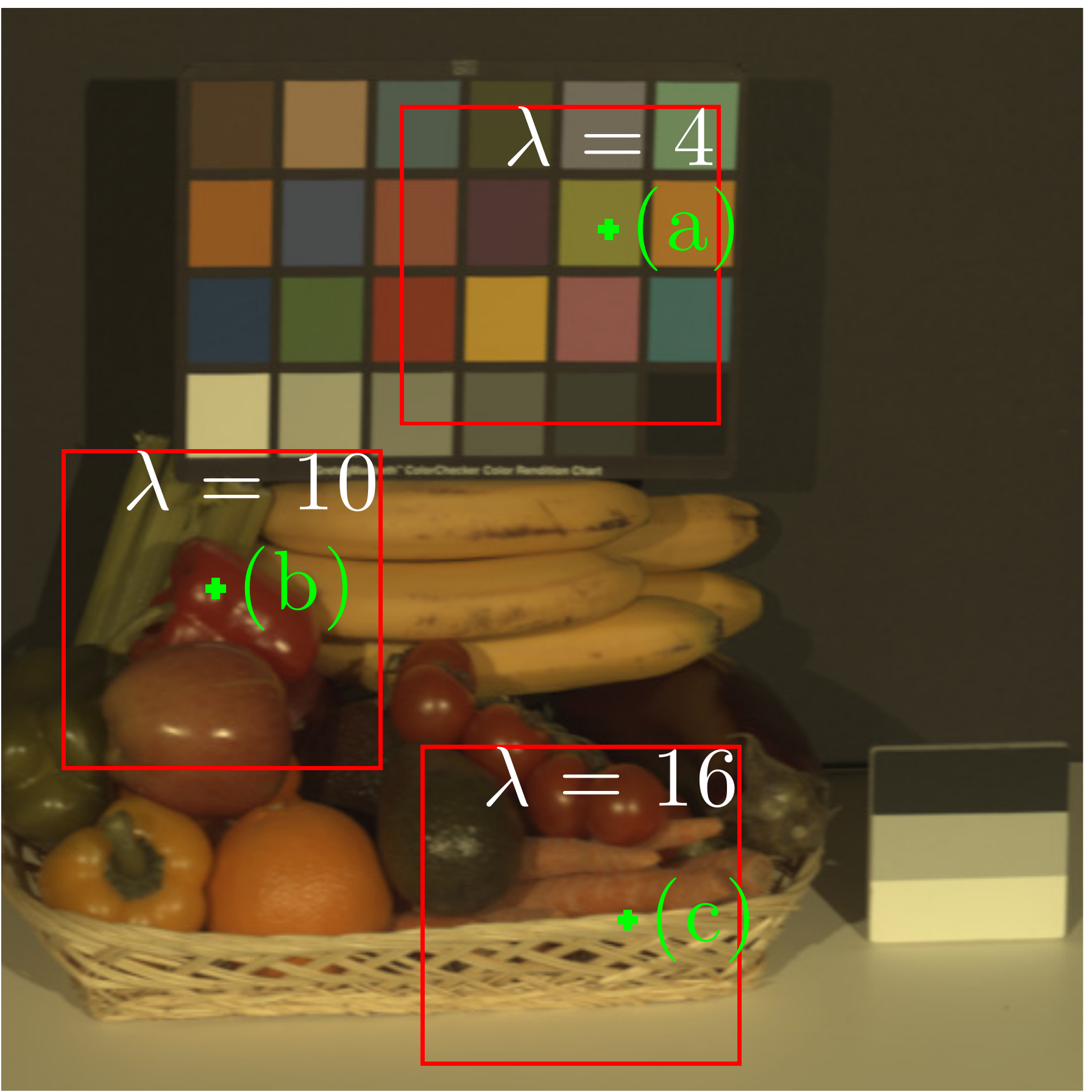}
  \includegraphics[width=4cm]{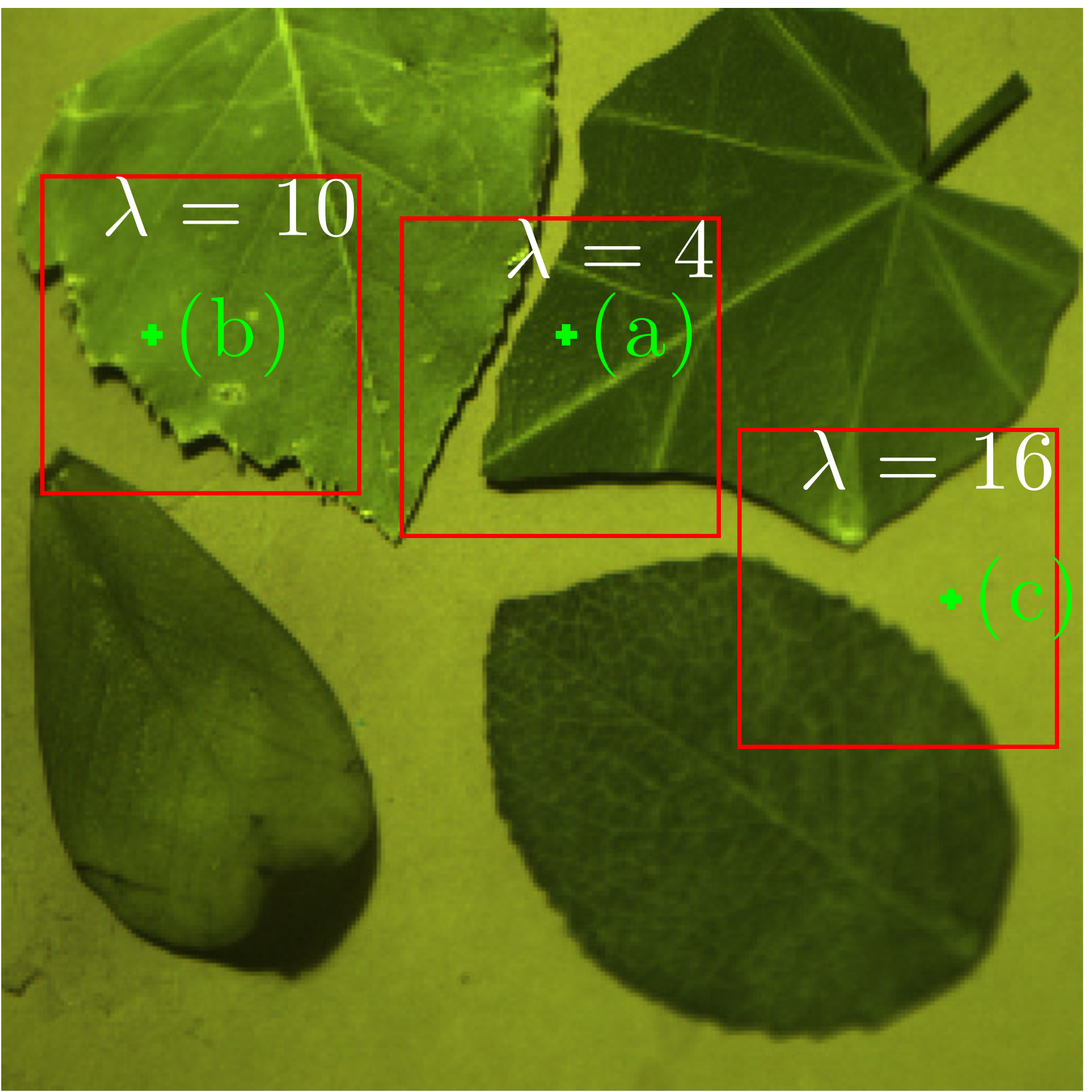}
\caption{\scriptsize Both reference HSI volumes (false RGB color). Left: Groundtruth for synthetic experiments (``\emph{Fresh Fruits}''). Right: Naive demosaicing of data from real sensor (``\emph{Leaves}'').}
\label{fig:groundtruth}
\end{figure}

This section demonstrates the capability of the proposed method. In the first part, we use a simulated model of the proposed architecture. The synthetic measurements are produced from an existing real high resolution HSI dataset (Fig.~\ref{fig:groundtruth}, left~\cite{Skauli2013}). This allows a quantitative comparison of the quality in several experimental conditions reproduced in a controlled numerical environment. The second part uses compressed data acquired with the existing \emph{mosaic} version of the sensor described below and in~\cite{geelen2014compact} (Fig.~\ref{fig:groundtruth}, right). For this imager, a naive demosaicing method is implemented in~\cite{geelen2014compact}. Notice that in all experiments, the number of iterations for PIHT was set to $S=200$.

\begin{figure}[t]
\newcommand{\fsp}{\hspace{0mm}}
\newlength{\figwda}\setlength{\figwda}{3cm}
\newlength{\figwdb}\setlength{\figwdb}{3.6cm}
\newlength{\figrb}\setlength{\figrb}{1mm}
\newlength{\figv}\setlength{\figv}{-3mm}
\centering
\begin{tabular}{@{\fsp}c@{\fsp}c@{\fsp}c@{\fsp}c@{\fsp}c@{\fsp}}
\includegraphics[width=\figwda]{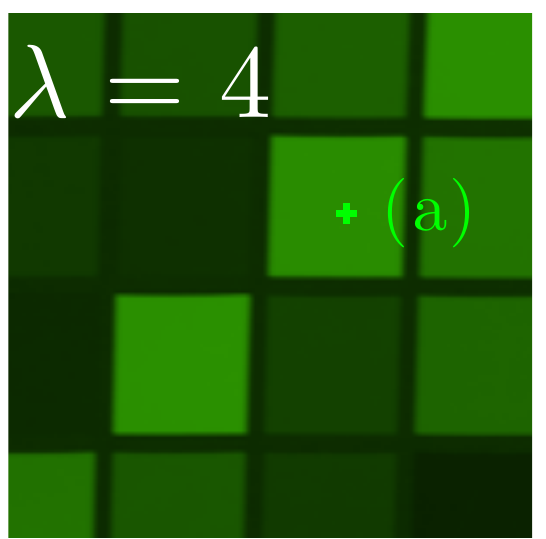}&
\includegraphics[width=\figwda]{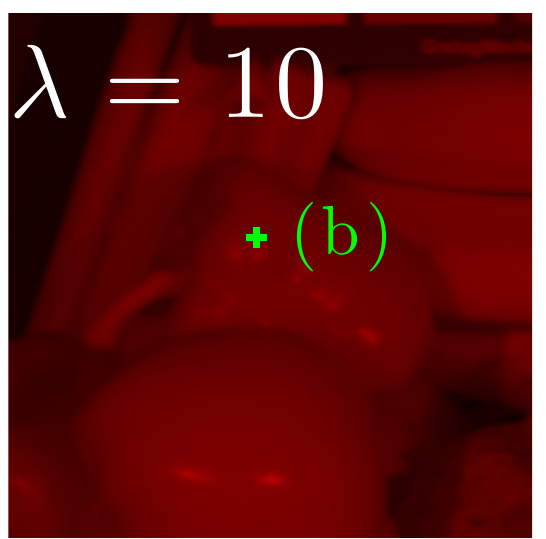}&
\includegraphics[width=\figwda]{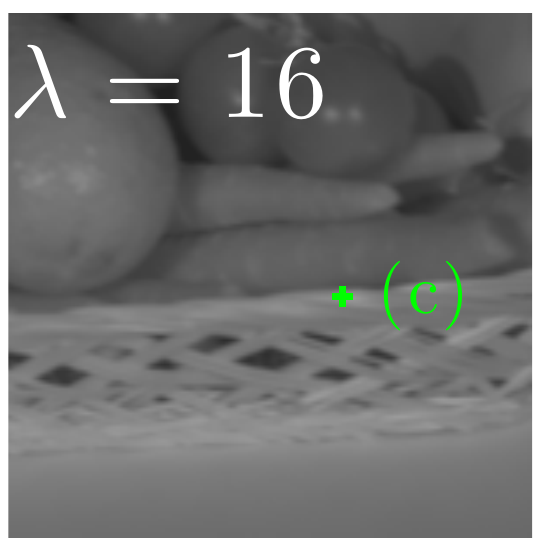}&
\raisebox{\figrb}{\includegraphics[width=\figwdb]{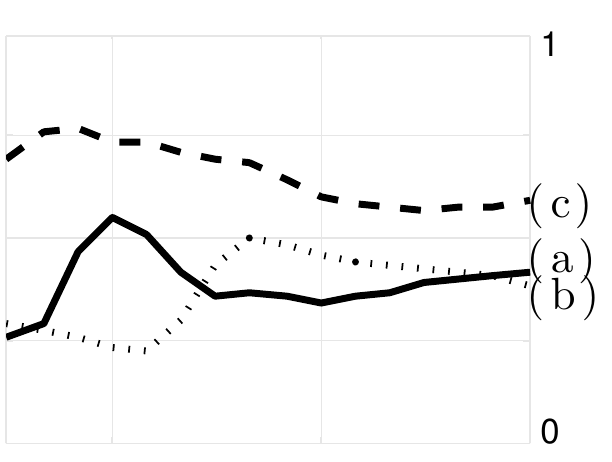}}\\[\figv]
\includegraphics[width=\figwda]{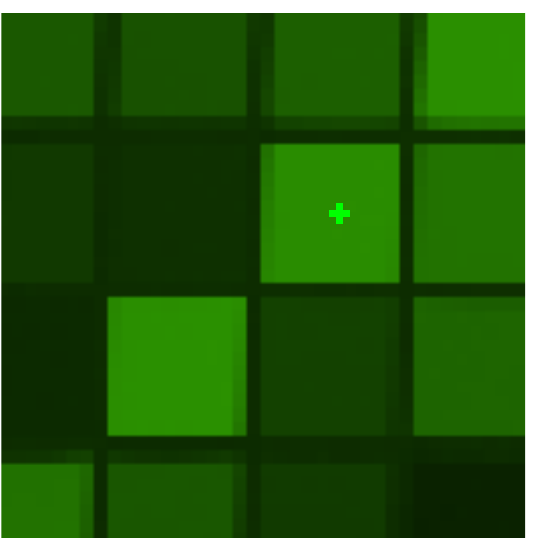}&
\includegraphics[width=\figwda]{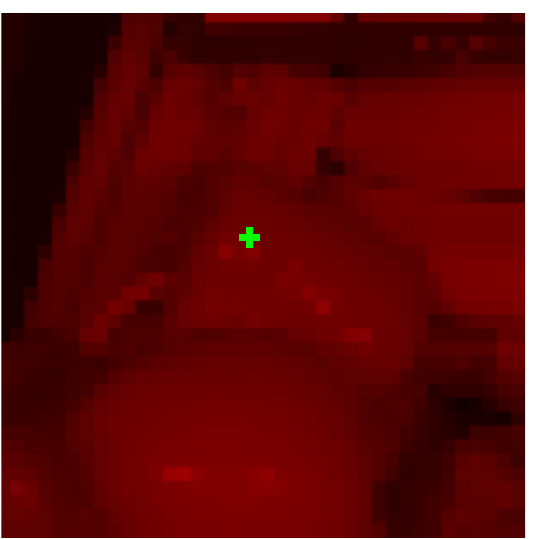}&
\includegraphics[width=\figwda]{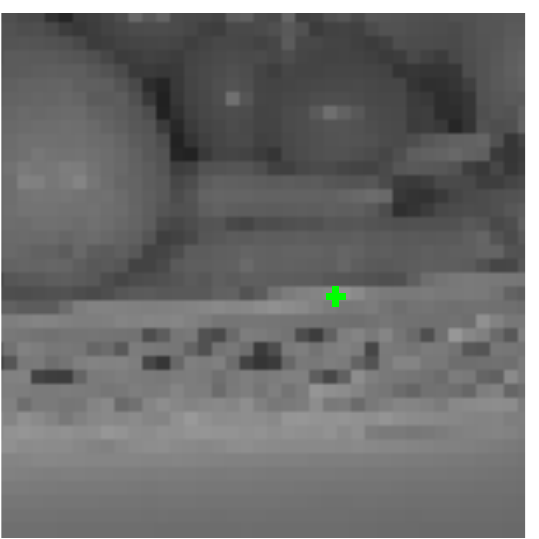}&
\raisebox{\figrb+0.25mm}{\includegraphics[width=\figwdb]{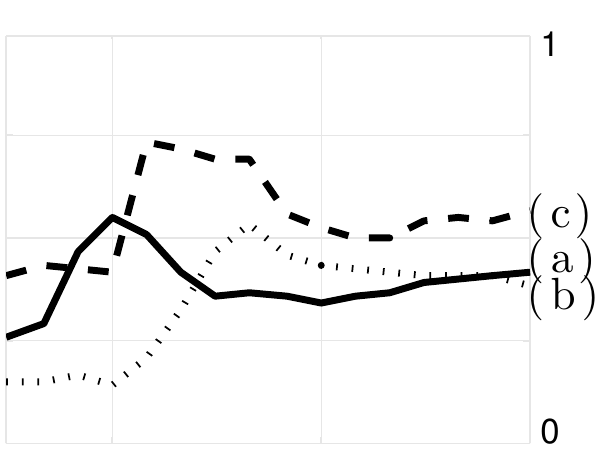}}\\[\figv]
\includegraphics[width=\figwda]{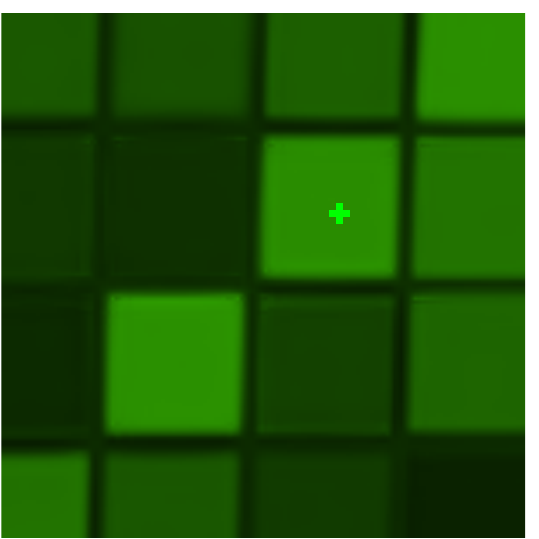}&
\includegraphics[width=\figwda]{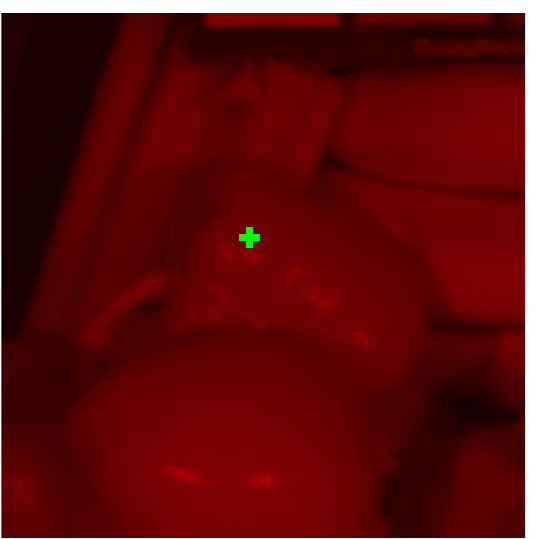}&
\includegraphics[width=\figwda]{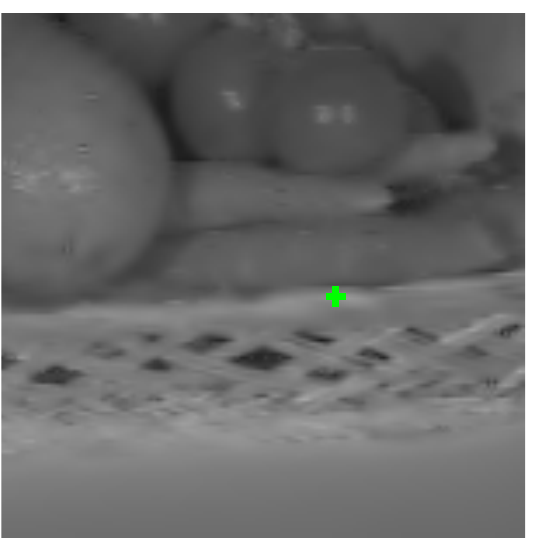}&
\raisebox{\figrb+0.25mm}{\includegraphics[width=\figwdb]{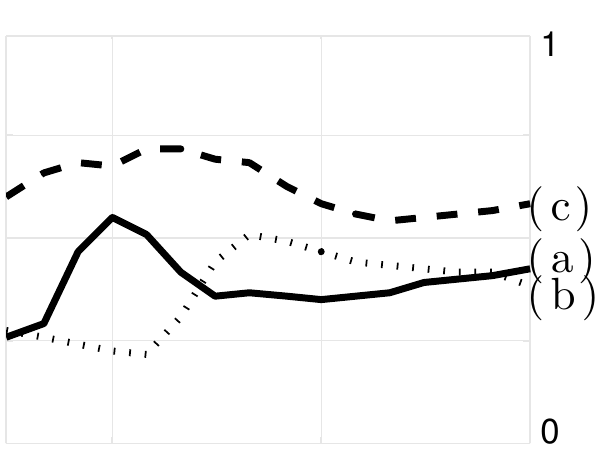}}\\[\figv]
\includegraphics[width=\figwda]{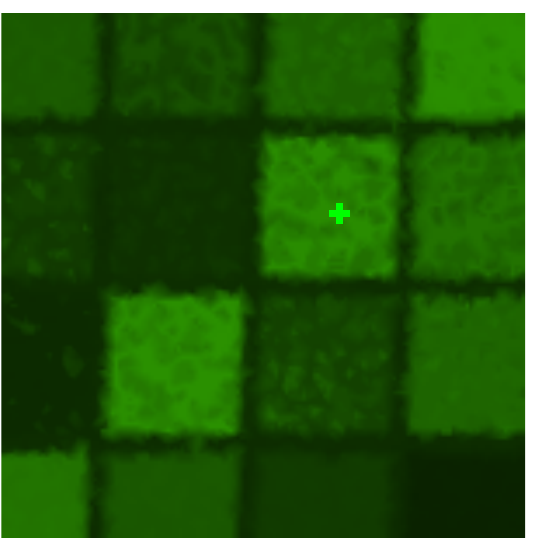}&
\includegraphics[width=\figwda]{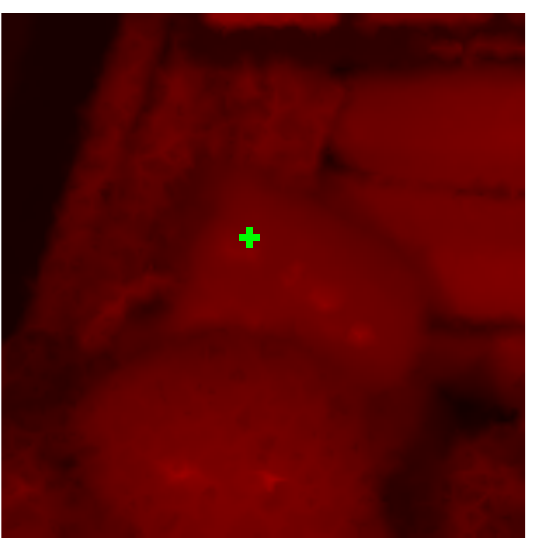}&
\includegraphics[width=\figwda]{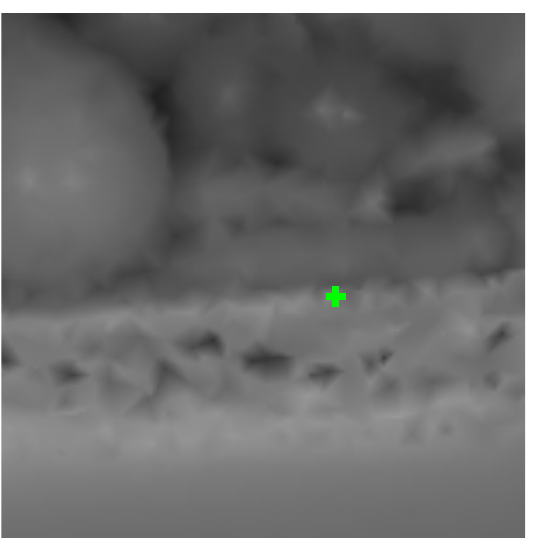}&
\raisebox{\figrb+0.25mm}{\includegraphics[width=\figwdb]{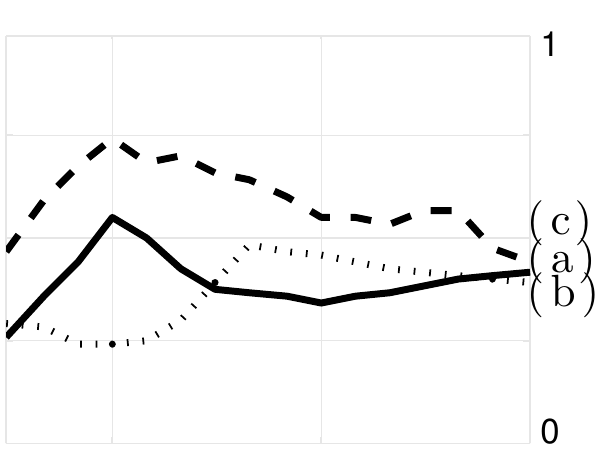}}\\[\figv]
\includegraphics[width=\figwda]{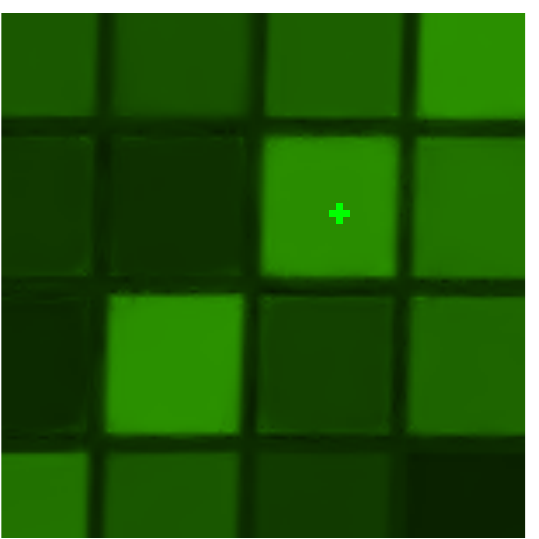}&
\includegraphics[width=\figwda]{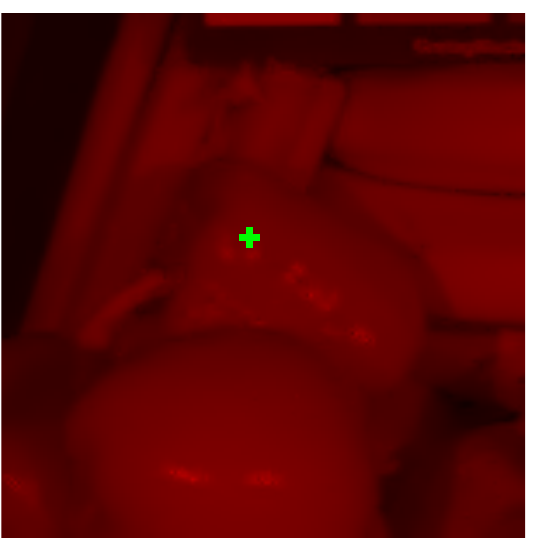}&
\includegraphics[width=\figwda]{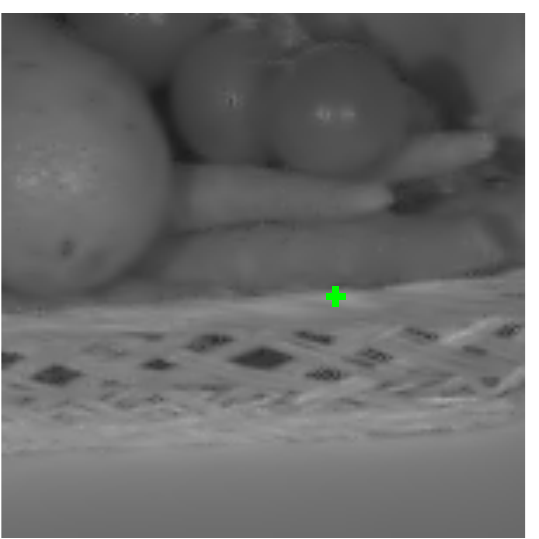}&
\raisebox{\figrb-1mm}{\includegraphics[width=\figwdb]{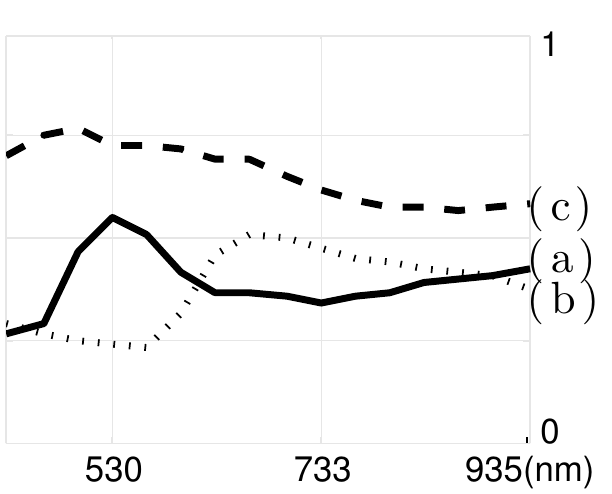}}
\end{tabular}
\caption{\scriptsize From top to bottom: Groundtruth; Naive demosaic 19.1 dB ; PIHT (mosaic) 27.03 dB; 3DLinterp (random) 23.5 dB; PIHT (random) 30.8 dB. Displayed bands are $530$nm (green, $\lambda=4$), $733$nm (deep red, $\lambda=10$) and $935$nm (infra red, $\lambda=16$).}
\label{fig:FRF}
\end{figure}


\medskip
\noindent{\bf Synthetic Data and Methods Comparison:} In this setup, the simulated sensor has a resolution $M$ of ${512\!\times\!512}$ pixels. The pixel grid is partitioned in $L\!=\!16$ different subsets corresponding to 16 FPI at wavelengths ranging between visible and near infra-red light. This paper proposes to test two different pixel subsets organizations: a mosaic of ${128\!\times\!128}$ identical ${4\!\times\!4}$ macro-pixels as for the existing sensor~\cite{geelen2014compact}, or a fully scrambled arrangement simply obtained by permuting randomly the pixel indices over the whole FPA, hence preserving the conditions on masks $\{\bs M_\lambda\}$ (Sec.~\ref{sec:HSInpainting}). In Table~\ref{SNRtable} we compare, for the random case, the 3-D linear interpolation (3DLinterp, also used as initialization of other methods), with a convex reconstruction method (ABPDN~\cite{Candes2011c} solved with the algorithm from~\cite{Antoine2014,Goldstein2013b,chambolle:2010}) and PIHT. Three target resolutions are tested: the Nyquist rate ${128\!\times\!128\!\times\!16}$; an intermediate case (available thanks to our generalized inpainting formulation) ${256\!\times\!256\!\times\!16}$; and a fully compressive (plain inpainting) ${512\!\times\!512\!\times\!16}$ setup. The convex method was stopped after 2000 iterations, possibly before full convergence. This probably explains its low reconstruction SNR levels. Speed was the main motivation for using a greedy method in this work but PIHT SNRs also appeared to clearly outperform those of APBDN in this setup. Note that when the super-resolution factor $N/M$ increases, while the reconstruction SNR degrades rapidly for 3DLinterp and ABPDN, it remains almost constant for PIHT. However, horizontal comparison in the table has to be interpreted with care since the SNR may depend on $N$.

On Fig.~\ref{fig:FRF}, we fix $N = 16M$ and compare the existing mosaic sensor (rows 2 and 3) with the random layout (4 and 5). We also compare PIHT (3 and 5) with a naive demosaicing (nearest neighbor interpolation on each band individually, row 2) and 3DLinterp method (row 4). Observe first that PIHT gives almost perfect visual quality even if the super-resolution factor $N/M$ is as high as 16. The advantage of the random pattern over the mosaic is best seen by looking at band 16 and at the spectrum (c) from the edge of the wicker basket. Both naive and 3DLinterp methods clearly show bad performances compared to PIHT. Finally, for the sake of completeness, we have added noise to the simulated compressive data and studied the average reconstruction SNR that was obtained with PIHT in the case $N=4M$. For an input SNR of $10$\,dB, $20$\,dB and $30$\,dB, we observed a reconstruction SNR of $18.9$\,dB, $27.0$\,dB and $32.3$\,dB, respectively, demonstrating the robustness of the method.

\begin{table}
\centering
\newcommand{\tbs}{\hspace{1mm}}
\begin{tabular}{|l@{\tbs}|@{\tbs}c@{\tbs}|@{\tbs}c@{\tbs}|@{\tbs}c@{\tbs}|@{\tbs}c@{\tbs}|@{\tbs}c@{\tbs}|@{\tbs}c@{\tbs}|}

\hline
 	& \multicolumn{2}{@{\tbs}c@{\tbs}|@{\tbs}}{$N = M$ } & \multicolumn{2}{@{\tbs}c@{\tbs}|@{\tbs}}{$N = 4M$} & \multicolumn{2}{@{\tbs}c@{\tbs}|}{ $N = 16M$}\\
\hline
	& SNR (dB) & time & SNR (dB) & time & SNR (dB) & time \\
\hline
3DLinterp & $28.3$ &  $34''$ & $25.3$ & $33''$ & $23.5$ & $33''$\\
ABPDN & $30.4$ & $59'29''$ & $17.4$ & $2$h$59'$ & $15.6$ & $12$h$52'$\\
PIHT & $32.9$ & $4'26''$ & $33.6$ & $15'6''$ & $30.9$ & $55'3''$ \\
\hline
\end{tabular}
\caption{Comparison of the initialization $\bx_0$ (3DLinterp) a convex method (ABPDN) and PIHT in terms of SNR (dB) and running time for different target sizes, given an acquisition size, on a synthetic example (generated from the ``Fresh Fruit" data set).}
\label{SNRtable}
\end{table}

\begin{figure}[t]
\newcommand{\afsp}{\hspace{0mm}}
\newlength{\afigwda}\setlength{\afigwda}{3cm}
\newlength{\afigwdb}\setlength{\afigwdb}{3.6cm}
\newlength{\afigrb}\setlength{\afigrb}{2mm}
\newlength{\afigv}\setlength{\afigv}{-3mm}
\centering
\begin{tabular}[t]{@{\afsp}c@{\afsp}c@{\afsp}c@{\afsp}c@{\afsp}c@{\afsp}}
\includegraphics[width=\afigwda]{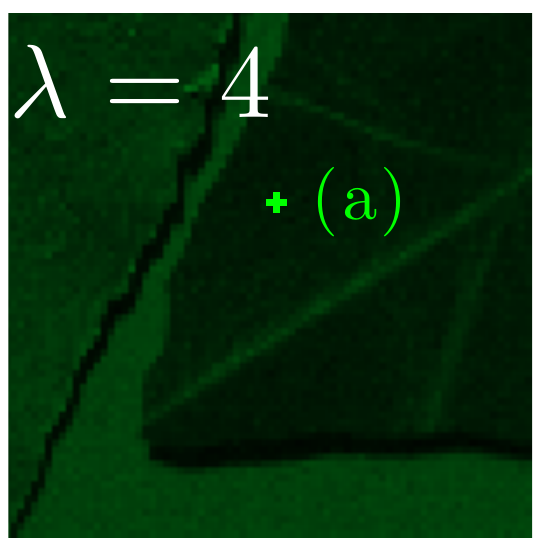}&
\includegraphics[width=\afigwda]{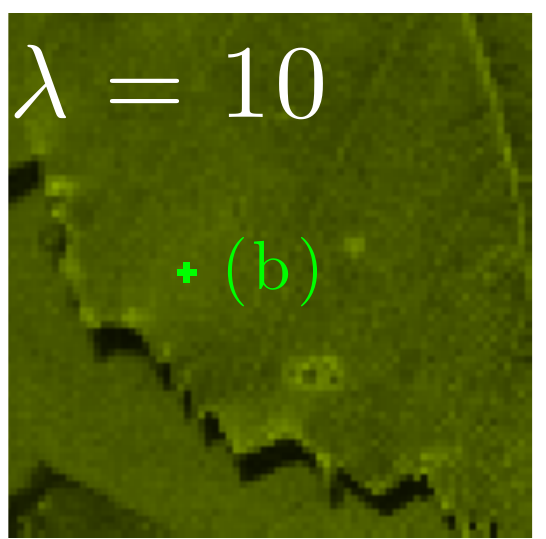}&
\includegraphics[width=\afigwda]{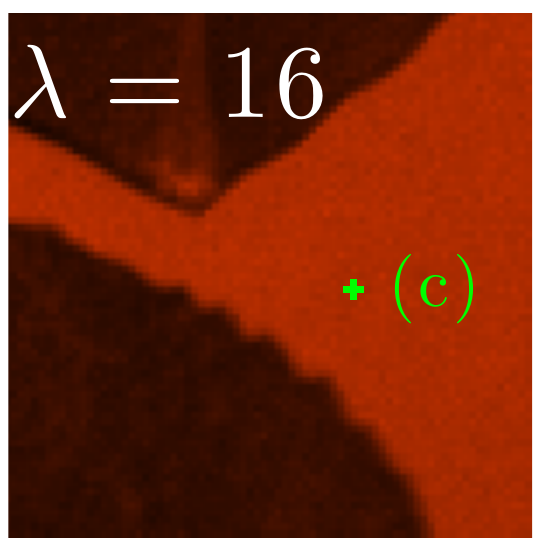}&
\raisebox{\afigrb}{\includegraphics[width=\afigwdb]{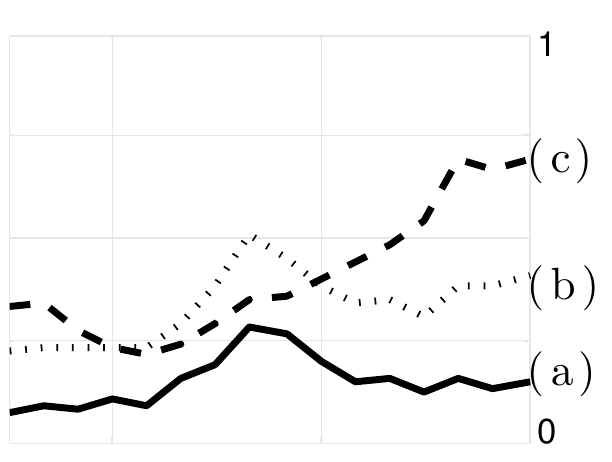}}\\[\afigv]
\includegraphics[width=\afigwda]{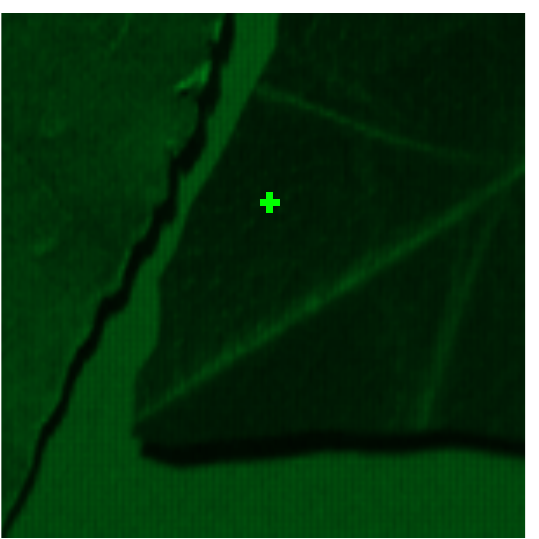}&
\includegraphics[width=\afigwda]{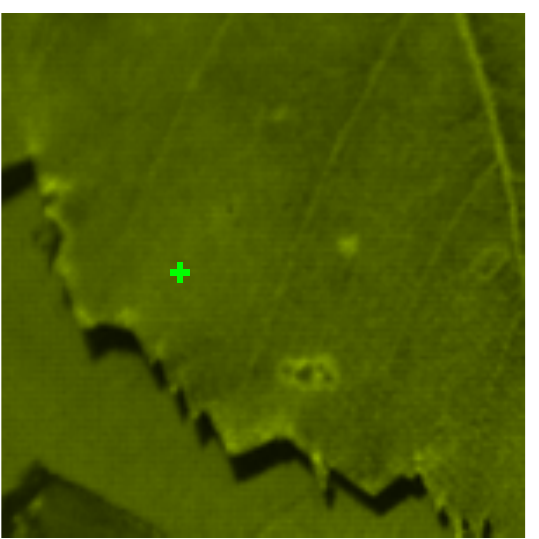}&
\includegraphics[width=\afigwda]{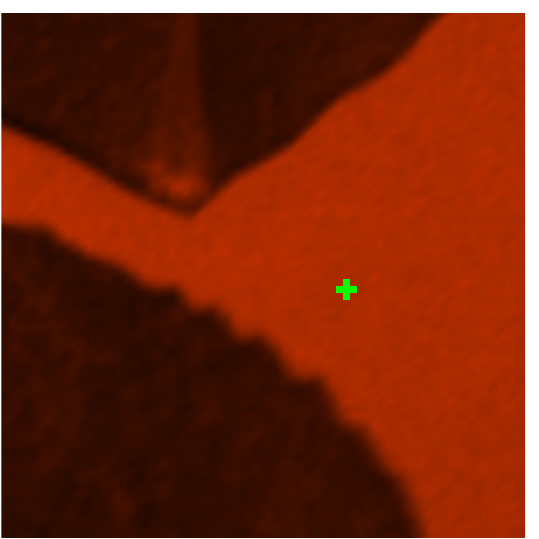}&
\raisebox{\afigrb+0.25mm}{\includegraphics[width=\afigwdb]{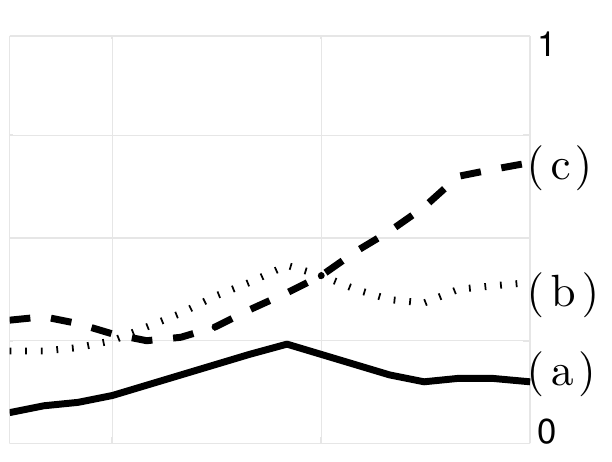}}\\[\afigv]
\includegraphics[width=\afigwda]{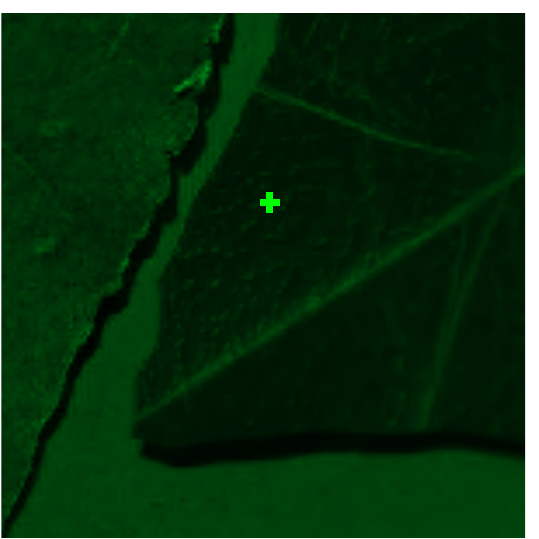}&
\includegraphics[width=\afigwda]{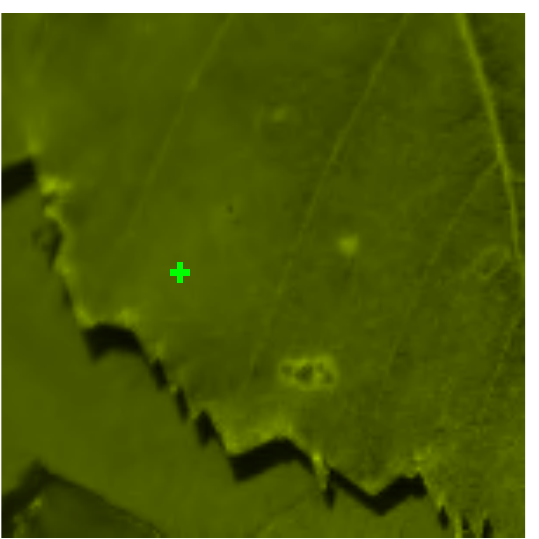}&
\includegraphics[width=\afigwda]{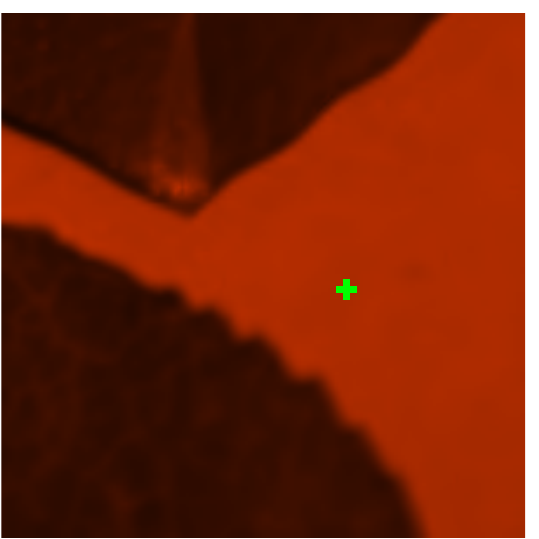}&
\raisebox{\afigrb-1mm}{\includegraphics[width=\afigwdb]{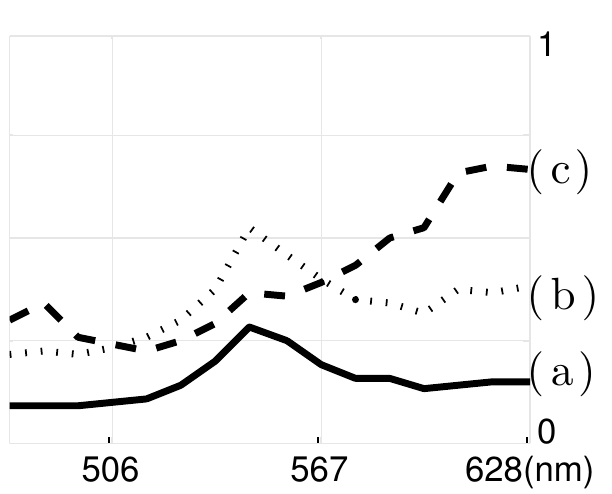}}
\end{tabular}
\caption{\scriptsize From top to bottom: Naive demosaic; 3DLinterp; PIHT. Displayed bands are $506$nm (bluish green, $\lambda=4$), $567$nm (green-yellow, $\lambda=10$) and $628$nm (red, $\lambda =16$).}
\label{fig:Leaves}
\end{figure}


\medskip
\noindent{\bf{Real Data Reconstruction:}} Here, the real sensor~\cite{geelen2014compact} was used. This imager has a resolution $M$ of $1024\!\times\!1024$ pixels organized in a mosaic of $256\!\times\!256$ identical $4\!\times\!4$ macro-pixels; each with $L = 16$ different FPI at wavelengths of visible light (as in Fig.~\ref{fig:FPA-FPI-explan}). This time, the results on Fig.~\ref{fig:Leaves} can only be compared qualitatively since no groundtruth is available. The super-resolution factor $N/M$ here is 4, \ie we recover a volume with $512\!\times\!512\!\times\!16$ voxels. The naive method that was used prior to our work (top row) is clearly the worst. Notice that the performances of 3DLinterp seems much better than in Fig.~\ref{fig:FRF}. However, in the electronic version of this article, you can zoom on the middle row of Fig.~\ref{fig:Leaves} and see a ``grid artifact'' that is almost completely removed with PIHT (bottom row). 

\section{Conclusion}

\label{sec:conclusion}
In this paper, a novel hyperspectral imaging technique based on the snapshot imager from~\cite{geelen2014compact} was presented. We proposed a flexible generalized inpainting formulation inspired by compressed sensing and aiming at demosaicing and increasing the resolution of the acquired data volume. The formulation permits to choose arbitrarily the targeted resolution, allowing a tradeoff between quality and size. Randomly scrambling the distribution layout of the Fabry-P\'erot interferometers helped to stabilize further the reconstruction method. We regularized the inpainting problem with a redundant dictionary analysis sparsity prior and solved it using PIHT, a fast greedy iterative algorithm. We finally demonstrated the performances of our method, both qualitatively and quantitatively, through numerical simulations on synthetic and real data. As in~\cite{geelen2014compact}, our method allows fast snapshot acquisition but with respect to the original imager, it dramatically increases the resulting resolution. The main shortcoming is that spectral resolution is limited to a few wavelengths. While it can suffice in many applications, others may require a finer spectral resolution. This is also a big difference with respect to other research in compressive HSI~\cite{Gehm2007}. Future works will include filter layout optimization or integration of true spectral filter responses for increased spectral resolution. Combining PIHT reconstruction with coded aperture compressive FP filtered HSI architectures \cite{Degraux:iTWIST14} is also envisioned. 

\section*{Acknowledgements}
\label{sec:acknowledgements}

KD and LJ thank G. Peyr\'e and J. Fadili for interesting discussions on the implementations of PIHT and its connection with oblique projectors. KD and LJ are funded by the F.R.S.-FNRS. Computational resources have been provided by the supercomputing facilities of the CISM/UCL and the C\'ECI in FWB funded by the F.R.S.-FNRS under convention 2.5020.11.

\end{document}